\documentclass[runningheads]{llncs}
\usepackage{graphicx}
\usepackage{multirow}
\usepackage{amsmath}
\usepackage{bbold}

\renewenvironment{quote}{%
  \list{}{%
    \leftmargin0.5cm   
    \rightmargin\leftmargin
  }
  \item\relax
}
{\endlist}

\usepackage{lineno} 

\begin{document}
\title{A Language Model based Framework for New Concept Placement in Ontologies}
\titlerunning{A Language Model based Framework for Ontology Concept Placement}

\author{Hang Dong\inst{1,2}\orcidID{0000-0001-6828-6891} \and
Jiaoyan Chen\inst{3,1}\orcidID{0000-0003-4643-6750} \and \\
Yuan He\inst{1}\orcidID{0000-0002-4486-1262} \and
Yongsheng Gao\inst{4}\orcidID{0000-0002-3468-2930} \and \\
Ian Horrocks\inst{1}\orcidID{0000-0002-2685-7462}}
\authorrunning{H. Dong et al.}
%
\institute{University of Oxford, Oxford, UK \\\email{\{yuan.he,ian.horrocks\}@cs.ox.ac.uk} \and University of Exeter, Exeter, UK \\\email{h.dong2@exeter.ac.uk} \and University of Manchester, Manchester, UK \\\email{jiaoyan.chen@manchester.ac.uk} \and SNOMED International, London, UK \\\email{yga@snomed.org}}
\maketitle              
\begin{abstract}
We investigate the task of inserting new concepts extracted from texts into an ontology using language models. We explore an approach with three steps: \textit{edge search} which is to find a set of candidate locations to insert (i.e., subsumptions between concepts), \textit{edge formation and enrichment} which leverages the ontological structure to produce and enhance the edge candidates, and \textit{edge selection} which eventually locates the edge to be placed into. In all steps, we propose to leverage neural methods, where we apply embedding-based methods and contrastive learning with Pre-trained Language Models (PLMs) such as  BERT for edge search, and adapt a BERT fine-tuning-based multi-label Edge-Cross-encoder, and Large Language Models (LLMs) such as GPT series, FLAN-T5, and Llama 2, for edge selection. We evaluate the methods on recent datasets created using the SNOMED CT ontology and the MedMentions entity linking benchmark. The best settings in our framework use fine-tuned PLM for search and a multi-label Cross-encoder for selection. Zero-shot prompting of LLMs is still not adequate for the task, and we propose explainable instruction tuning of LLMs for improved performance. Our study shows the advantages of PLMs and highlights the encouraging performance of LLMs that motivates future studies.

\end{abstract}
%
%
%
\section{Introduction}

New concepts appear as they are discovered in the real world, for example, new diseases, species, events, etc. Ontologies are inherently incomplete and require evolution by enriching with new concepts. A main source for concepts is corpora, e.g., new publications that contain mentions of concepts not in an ontology.

In this work, we focus on the problem of placing a new concept into an ontology by inserting it into an edge which corresponds to a subsumption relationship between two atomic concepts, or between one atomic concept and one complex concept constructed with logical operators like existential restriction ($\exists r.C$). Distinct from previous work in taxonomy completion (e.g., \cite{Wang2022QEN,Zeng2021,zhang2021TMN}), the task allows natural language contexts together with the mention as an input and also considers logically complex concepts by Web Ontology Language (OWL). Distinct from previous work in new entity discovery (e.g., \cite{dong2023NIL}), the task places the new entity into the ontology, a step further to their discovery from the texts. The task is more challenging than entity linking from a mention to a concept, considering that there are many more edges than already the large number of concepts and axioms in an ontology (of a form much more complex than a tree), even by limiting the edges to only those having one-hop or two-hop. 

Recently, machine learning, neural network based methods, and especially pre-trained language models (PLM), have been applied to ontology engineering tasks. For new entity discovery tasks, typically, the entity linking or retrieval tasks comprise two steps, the first is to search relevant entities by narrowing down the candidates, and the second is to select the correct one. Previous studies on entity linking and new entity discovery mostly use BERT-based fine-tuning methods \cite{wu2020blink,dong2023NIL}. We differ Large Language Models (LLMs) from PLMs by their vast difference in scale and language generation capabilities. There is a recent growth of studies using LLMs, e.g., for entity linking \cite{wang2023llm-el} and ontology matching \cite{he2023llm-om}, but the experimental results are yet to be confirmed and their advantages and drawbacks for concept placement are not clear. A more detailed investigation is needed to compare the methods for the representation, and a framework is needed for their comparison. In the texts below, we use LMs as a general term for both PLMs and LLMs and use more specific terms where necessary. 

For concept placement, we propose a framework that extends the two-step process, with another edge enrichment step. After the edge search to narrow the edge candidates to a limited number, we enrich the edges by walking in the ontological graph by extending the parents and children to another layer. Then this enriched set of edges is re-ranked through the edge selection part (which can be modelled as a multi-label classification task). Using this framework, we are able to compare different data representation methods, including traditional inverted index, fixed embedding based similarity, contrastive learning based PLM fine-tuning, and instruction-tuning and prompting of LLMs.

The evaluation is based on the recent datasets in \cite{dong2023oet}, created by using an ontology versioning strategy (i.e., comparing two versions of an evolving ontology) to synthesise new concepts and their gold edges to be placed w.r.t. the older version of the ontology. The ontology is SNOMED CT, under \textit{Disease} and \textit{CPP} (Clinical Finding, Procedure, and Pharmaceutical / biologic products) branches. 

Results indicate that edge enrichment by leveraging the structure of ontology greatly improves the performance of new concept placement. Also, among the data representation methods, contrastive learning based PLM fine-tuning generally performed the best in all settings. The inadequate yet encouraging results of LLMs under our experimental setting may be related to the input length restriction and the inherent knowledge deficiencies of LLMs for nuanced concept relations of domain specific ontologies. Instruction-tuning, especially with automated explainable prompts, improves over the zero-shot prompting (i.e., no further instruction-tuning) of LLMs. Our results suggest the potential of LLMs and motivate future studies to leverage them for ontology concept placement.\footnote{
Our implementation of the methods and experiments are available at \url{https://github.com/KRR-Oxford/LM-ontology-concept-placement}.}

\section{Related Work}

\subsection{Ontology Concept Placement}

Ontology concept placement is a key task in ontology engineering and evolution. It aims to automatically place or insert a new concept, in its natural language form and potentially with contexts in a corpora, to an existing ontology. This automated task helps to reduce the immense initial human effort to discover and insert new concepts, as humans may not be able to review all available new information at the rate when they are available, and the manual process while of high quality, is of high cost and low efficiency \cite{glauer2022interpretable,tgdk2023lifesci}. 

The recent study in \cite{dong2023oet} summarised the related available datasets on ontology concept placement. Datasets for the relevant tasks include taxonomy completion, ontology extension, post-coordination, and new mention and entity discovery. The proposed new datasets in \cite{dong2023oet} supports a more comprehensive set of characteristics, including NIL entity discovery, contextual mentions, concept placement (under both atomic and complex concepts in ontologies). We extend the datasets in the work \cite{dong2023oet} and use them for benchmarking in this paper.

Another relevant task is entity linking, which links a textual mention to its concept in a Knowledge Base (KB) or an ontology \cite{shen2014entity}. Entity linking can be extended to the case for out-of-KB mentions \cite{dong2023NIL}. Ontology Concept Placement is distinct from entity linking to a concept, which alternatively links an out-of-KB mention to an edge (of subsumption relations) in the structure of an ontology. 

\subsection{Pre-trained Language Models for Ontology Concept Placement}

We consider pre-trained language model as a neural, Transformer model \cite{Vaswani2017Transformers} that can be pre-trained using corpora using masked modelling or by predicting future tokens, processing very large amounts of text \cite{jurafsky2023slpbook}. A Large Language Model is a scaled PLM to a vastly higher degree which can result in improved performance and emergent capabilities \cite{zhao2023survey}.

A relevant line of work to ontology concept placement is Knowledge Graph Construction, where BERT is evaluated and shows promise to enhance several relations in WikiData \cite{Veseli2023}. Other studies focus on formal KBs which are usually expressed as OWL Ontology, e.g., by predicting the subsumption relations \cite{chen2023bertsub}. The work \cite{hertling2023transformer} predicts a wider range of inter-ontology relations (e.g., equivalence, subsumption, meronymy, etc.) using PLMs (e.g., DistillBERT, RoBERTa, etc.).

For ontology concept placement, PLMs have also been applied. The study \cite{Liu2020placement} aims to place concepts to SNOMED CT by pre-training and fine-tuning BERT for subsumption prediction. The study \cite{ruas2022} uses a similar BERT-based Bi-encoder architecture and experiments with more medical ontologies. However, both works always place a concept as a leaf node, instead of higher levels. 

Another approach utilising LLMs is through a prompting-based approach. The idea is to formulate an ontology-related task using natural language input that leverages the generative capability of a language model. While the recent study \cite{wang2023llm-el}  explored prompting-based approaches of LLMs for concept equivalence linking, few studies have explored them for ontology concept placement.

In this paper, we propose an LM-based framework that leverages embedding, fine-tuning, prompting, and instruction-tuning of PLMs and LLMs for ontology concept placement. The task also considers contexts in a mention and the logically complex concepts in ontologies that are not considered in previous work. 

\section{Problem Statement}

We use the definition of an OWL ontology, a Description Logic KB that contains a set of axioms \cite{GRAU2008OWL2,baader_horrocks_lutz_sattler_2017chap8}. We focus on the TBox (terminology) part of an ontology, containing General Concept Inclusion axioms, each as $A \sqsubseteq B$, where $A$ (and $B$) are atomic or complex concepts \cite{baader_horrocks_lutz_sattler_2017chap2}. \textit{Complex concepts} mean concepts that involve at least one logical operator, e.g., negation ($\neg$), conjunction ($\sqcap$), disjunction ($\sqcup$), existential restriction ($\exists r.C$), universal restriction ($\forall r. C$), etc. \cite{baader_horrocks_lutz_sattler_2017chap2}. 

An ontology $\mathcal{O}$ can be more simply defined as a set of concepts $D$ (possibly complex) and directed edges $E$. A directed edge contains a direct parent and a direct child, where the parent or child can be complex concepts.\footnote{We focus on the common case that only the parent can be a complex concept, as in the explicit axioms in the SNOMED CT ontology.}

Formally, the task is to place a new concept mention $m$ (with surrounding contexts in a corpus) into edges in an ontology $\mathcal{O}$ so that $C \sqsubseteq m \sqsubseteq P$ for an edge $<P,C>$ (or as $P \rightarrow C$) that contains a parent concept $P$ and a child concept $C$. The child concept $C$ can be \texttt{NULL} when the mention is to be placed as a leaf node. Using SNOMED CT (version 1703) as an example, a mention ``Psoriatic arthritis'' (in a scientific paper) is to be placed as $\texttt{Psoriatic arthritis with distal interphalangeal joint involvement} \sqsubseteq \texttt{Psoriatic arthritis} \sqsubseteq \texttt{Psoriasis with arthropathy}$; and a mention ``Neurocognitive Impairment'' is to be placed as a leaf concept (so $C$ is \texttt{NULL}), and the axioms include $\texttt{Neurocognitive Impairment} \sqsubseteq \texttt{Cognitive disorder}$ and $\texttt{Neurocognitive Impairment} \sqsubseteq \exists \texttt{RoleGroup}.(\exists \texttt{DueTo}.\texttt{Disease})$\footnote{This means that Neurocognitive Impairment belongs to the role group \cite{spackman2002role} or a grouping of the characteristic that is caused by (``due to'') a disease.}.

Ontology concept placement can thus be considered matching from a textual mention (possibly surrounded by a context window) to edges in the structure of an ontology. Given that a concept may have more than one parent and more than one child, it can be placed into many edges. Thus ontology concept placement can be formulated as a \textit{multi-label learning} problem \cite{Gibaja2015multilabel}. The task is to learn a mapping function $f$ that can map the input (a textual mention possibly with contexts) to a set of labels (here as edges in $E$). Typically, a multi-label learning process can create a label \textit{ranking} based on a metric score that orders the whole set of labels \cite{Gibaja2015multilabel} or an ordered set without an explicit metric (e.g., by the order of text generation). This is distinct from the entity linking task which usually maps the input to only a single label (as an entity or a NIL entity) \cite{dong2023NIL}.

\section{Methodology}

Extending the general ideas in information retrieval and entity linking, we propose a three-step framework for ontology concept placement, as shown in Figure \ref{LM-framework} below. Usually, retrieving a set of correct items (e.g., edges) needs two steps, \textit{search} (or candidate generation) and \textit{selection} (or candidate ranking). The search step aims to find a set of seed concepts (to form edges) or a set of seed edges directly. The selection step finds (and also ranks, as in multi-label classification) the correct edges among the candidates. Considering the structural nature of the edge generation process, we add another step in between, \textit{edge formation and enrichment}, which forms seed edges from a seed concept (optionally) and enriches seed edges to derive the full candidate edges. We employ LMs in both the search and the selection steps, and further leverage the ontological structure for the edge formation and enrichment step.

\begin{figure}[t]
  \centering
  \includegraphics[width=0.9\textwidth]{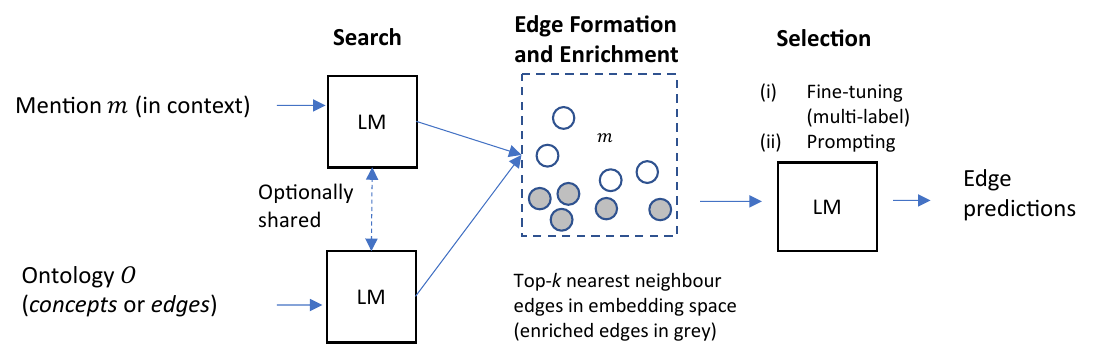}
  \caption{An overall three-step framework for ontology concept placement with LMs.}\label{LM-framework} 
\end{figure}

\subsection{Edge Search: Searching Seed Concepts or Edges}

The search step inputs a textual mention $m$ (with a context window) and an ontology $\mathcal{O}$, both represented using LMs. For concept search, we encode a mention and the label of a concept using an LM with fixed parameters (or as two same LMs sharing parameters). For edge search, we encode a mention and an edge using two LMs with fine-tuning to align them into the same embedding space, given the distinct types of texts (in corpora and in ontologies) between them.

\subsubsection{Concept Search with Fixed Embeddings} We search concepts by using the nearest neighbours of LM-based embeddings, i.e., ranking using the cosine similarity of the mention embedding and every concept embedding in the ontology. A domain-specific ontology-pre-trained BERT, SapBERT \cite{liu2021sap}, is used to represent both a mention and a concept. Complex concepts, with logical operators, can be verbalised using a rule-based verbaliser (e.g., in \cite{he-etal-2023-language}), before their embedding.

\subsubsection{Edge Search with Fine-tuning Edge-Bi-encoder} We use two LMs to encode the mention and the edge separately, using the representation of the [CLS] token in the last layer, adapting the Bi-encoder architecture \cite{wu2020blink,dong2023NIL}. A mention is represented as \texttt{[CLS] ctxt$_l$ [M$_s$] mention [M$_e$] ctxt$_r$ [SEP]}, where \texttt{ctxt$_l$} and \texttt{ctxt$_r$} are the left and right contexts of the mention in the document, resp., and \texttt{[M$_s$]}, \texttt{[M$_e$]} are the special tokens placed before and after the mention. In the setting without contexts, we set both \texttt{ctxt$_l$} and \texttt{ctxt$_r$} as empty strings. A directed edge (having a direct parent and a direct child) is represented as ``\texttt{[CLS] parent tokens [P-TAG] child tokens [C-TAG] [SEP]}''. We use a special token \texttt{[NULL]} to represent the \texttt{child tokens} of a leaf concept in the ontology.

The training follows a contrastive loss, more specifically, a max-margin triplet loss \cite{reimers-gurevych-2019-sentence} described below, where $\alpha$ is a margin of small value (e.g., 0.2) and $[x]_+$ denotes $\text{max}(x,0)$, for each mention to its gold edge (the $i$-th) in a batch, $s(m,e)$ is the mention-edge similarity, calculated as the dot-product of the mention embedding and the concept embedding. The idea is to make each mention close to one of its edges in the embedding space, but far away from the other edges within the same batch. We use in-KB data for training and validation to form a model and then finally validate and test on out-of-KB data. 
\begin{equation}\label{loss-bi-and-dot-product-score}
    L_{m_i,e_i} = \sum_{j \neq i}{[\alpha-s(m_i,e_{i})+s(m_i,e_{j})]_{+}}; \hspace{3mm}
    s(m,e) = v_m \cdot v_e
\end{equation} 

\subsection{Edge Formation and Enrichment}

The idea of edge formation and enrichment is to leverage the ontological structure together with the LM-based embedding for candidate retrieval. The detailed process with examples is presented in Figure \ref{Fig:Edge-enrich}.

\subsubsection{Edge Formation from Seed Concepts} 
\begin{figure}[t]
\centering
  \includegraphics[width=0.85\textwidth]{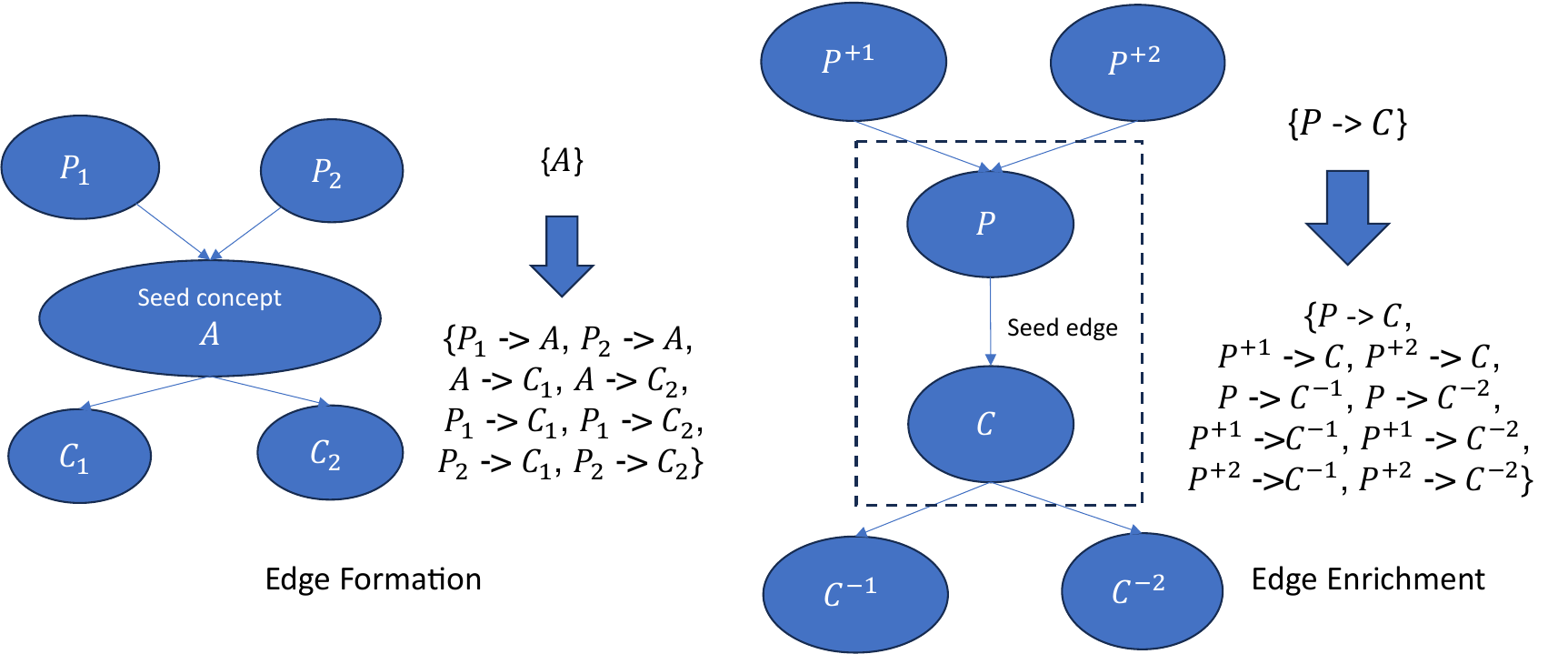}
  \caption{An example of the edge formation and enrichment process using ontology structure. Edge formation transforms a seed concept into a set of edges, while edge enrichment augments the set of edges one by one. For methods that directly search edges (e.g., Edge-Bi-encoder), no edge formation is needed and only enrichment is applied.}\label{Fig:Edge-enrich}
  
\end{figure}

When concept candidates are selected from entities, for each concept $A$, we traverse the ontology by one hop to find the parents $P_1,...,P_n$ and children $C_1,...,C_n$ of the concept, and then using the set  $S = \bigcup_i \{P_i\rightarrow A\} \cup \bigcup_j \{A\rightarrow C_j\} \cup \bigcup_i \bigcup_j \{P_i\rightarrow C_j\}$ as the candidate edge set, which includes all one-hop edges containing $A$ and all two-hop edges which traverse through $A$ (see an example in the left part of Figure \ref{Fig:Edge-enrich}). We further added leaf edges, $A \rightarrow \texttt{NULL}$,  to $S$.

\textbf{Edge Ranking after Edge Formation} Then the edge set is ranked using the LM-based embedding w.r.t. the mention $m$, as the average cosine similarity of $m$ to the parent and $m$ to the child in the embedding space (see Equation \ref{edge-emb-score-fixed} below). For the edge score of leaf edges (where $C = \texttt{NULL}$), we first set a rule to deduce whether the mention is to be placed on a leaf edge by checking if the top ranked seed concept is a leaf concept, if so, we prioritise all enriched leaf edges of the mention with the highest edge score (i.e., better ranked than non-leaf edges). 
\begin{equation}\label{edge-emb-score-fixed}
    \text{Edge\_score}_{fixed}(m, <P,C>) = \frac{\text{sim}(m,P) + \text{sim}(m,C)}{2} , \text{where } C \neq \texttt{NULL}
    \vspace{-3.5mm}
\end{equation}

\subsubsection{Edge Enrichment from Seed Edges} We further enrich the edges by traversing one-hop upper for parents and one-hop lower for children in the ontology. For each edge $P\rightarrow C$, we thus first find their one-hop upper parents $P^{+1}$,...,$P^{+i}$ and  one-hop lower children $C^{-1}$,...,$C^{-j}$, and enrich the set to $\{P\rightarrow C\} \cup \bigcup_i \{P^{+i}\rightarrow C\} \cup \bigcup_j \{P\rightarrow C^{-j}\} \cup \bigcup_i \bigcup_j \{P^{+i}\rightarrow C^{-j}\}$. We combine the enriched edges from all seed edges and then remove the duplicated edges (given that some of the enriched edges can be the same for different but similar seed edges). This covers more related edges based on the ontological structure and LM-based similarity and can greatly improve the recall of the edge retrieval. We also enrich a ``leaf'' edge, i.e., $P\rightarrow \texttt{NULL}$, when a parent $P$ in a non-leaf edge is predicted. 

\textbf{Edge Ranking after Edge Enrichment} The enriched edges are then ranked with scores from different edge search methods. For the fixed embedding approach, edges are ranked based on the edge score in Equation \ref{edge-emb-score-fixed}. For the fine-tuned embedding (Edge-Bi-encoder) approach, edges are ranked using the dot product scores, $s(m,e)$ (see Equation \ref{edge-emb-score-fine-tuned} below and the right part of Equation \ref{loss-bi-and-dot-product-score}) for all edges (including both leaf and non-leaf edges) after the enrichment for the fine-tuned embedding (Edge-Bi-encoder) approach. The top-$k$ candidate edges are then retrieved from the seed edges after this process. 
\begin{equation}\label{edge-emb-score-fine-tuned}
    \text{Edge\_score}_{fine-tuned}(m, <P,C>) = s(m, <P,C>)
\end{equation}

\subsection{Edge Selection}
\label{sec:edge-select}

The edge selection step aims to find the correct edges to place the concept mention from the $k$ candidate edges. We utilise LMs based on their distinct architectures, i.e., we fine-tune BERT-like, encoder-only PLMs for multi-label classification, and prompts and instruction-tunes LLMs, which have a decoder, for result generation.

\subsubsection{Fine-tuning PLMs: Multi-label Edge-Cross-encoder} 

We adapt an LM-based cross-encoder in \cite{dong2023NIL,wu2020blink}, that encodes the interaction between sub-tokens in the contextual mention and an edge in the top-$k$ edges, for multi-label classification. Specifically, for each of the $k$ candidate edges the input is a concatenation of the contextual mention with the edge, i.e., \texttt{[CLS] ctxt$_l$ [M$_s$] mention [M$_e$] ctxt$_r$ [SEP] parent tokens [P-TAG] child tokens [C-TAG] [SEP]}, and the output is a multi-label classification over all the inputs, i.e., the selection of the correct edges from the candidates. We use a special token \texttt{[NULL]} to represent the \texttt{child tokens} of a leaf edge. Each input is encoded with a BERT model into a vector $v_{cross}$ (we use the representation of \texttt{[CLS]} in the last layer).

Therefore, the loss is a binary cross-entropy loss after a sigmoid activation of the score, linearly transformed from the representation vector, $s^{(cross)}_{m,e}=v_{cross} w$, of each input. All the inputs share the same BERT model for fine-tuning.

\subsubsection{Zero-shot Prompting LLMs} Alternatively, a recent paradigm is to prompt LLMs to generate answers directly. We formulate a prompt to allow LLMs to generate the indices of the options. The prompt provides contexts and all necessary information including the top-$k$ candidate edge options to allow the LLMs to be conditioned and generate the answer. The prompt is structured as below, which contains an input (including task description, the mention in context, and the options of $k$ edges) and a response headline. The sequence which is underlined (after \#\#\# Response) is expected to be generated by the LLM.

\begin{quote}
\scriptsize
    \#\#\# Input:
    
    Can you identify the correct ontological edges for the given mention (marked with *) based on the context? The ontological edge consists of a pair where the left concept represents the parent of the mention, and the right concept represents the child of the mention. If the mention is a leaf node, the right side of the edges will be \texttt{NULL}. If the context is not relevant to the options, make your decision solely based on the mention itself. There may be multiple correct options. Please answer briefly using option numbers, separated by commas. If none of the options is correct, please answer None.
    
    \vspace{1mm}
    mention in context:
    
    Our aim was to verify the occurrence of selected mutations of the EZH2 and ZFX genes in an Italian cohort of 23 sporadic *parathyroid carcinomas*, 12 atypical and 45 typical adenomas.
    
    \vspace{1mm}
    options:
    
    0.primary malignant neoplasm $\rightarrow$ parathyroid carcinoma
    
    1.malignant neoplastic disease $\rightarrow$ malignant tumor of parathyroid gland
    
    2.malignant neoplastic disease $\rightarrow$ primary malignant neoplasm of parathyroid gland

    ...
    
    8.primary malignant neoplasm of parathyroid gland $\rightarrow$ \texttt{NULL}

    ...
    
    (till all the $k$ candidates are listed)
    
    \vspace{1mm}
    \#\#\# Response:
    
    \underline{2,8}
\end{quote}

\subsubsection{Explainable Instruction-tuning LLMs} We can observe that it would not be straightforward for an LLM to directly figure out the edges and output the option numbers (e.g., 2,8 in the example above). Fine-tuning with in-KB training data would be needed. To bridge the reasoning gap between the input and the response, we propose to add an explanation section that describes the reasoning steps in a narrative form. 

Thus, we automatically synthesise explanations by steps to solve the new concept placement problem: (i) List all possible parents in the candidates; (ii) Find correct parents; (iii) Narrow the list of children based on the correct parents; (iv) Find correct children; (v) List the final answer based on the correct children.  The explanation \texttt{Expl\_texts} is a function of the $k$ candidate edges and the gold edges of a mention, i.e., $\texttt{Expl\_texts} = \text{Template}(E_{cand}, E_{gold})$. 
The template is below, where elements in the lists (in square brackets) are separated by comma.

\begin{quote}
\scriptsize
    \#\#\# Explanation:

    From the parents in the options above, including \texttt{[all candidate parents]}, the correct parents of the mention, \texttt{[mention name]}, include \texttt{[correct gold parents]}. Thus the options are narrowed down to \texttt{[option numbers having correct gold parents]}. From the children in the narrowed options, including \texttt{[children in the filtered options]}, the correct children of the mention, \texttt{[mention name]}, include \texttt{[correct gold children in the filtered options]}. Thus, the final answers are \texttt{[correct option numbers]}.
    
\end{quote}

We place the explanation section (\#\#\# Explanation) before the response section (\#\#\# Response). During training, the whole explanation is fed into the LLM to allow it to be conditioned to generate the response. During inference, the instruction-tuned LLM is expected to generate an explanation of the same template structure, after the explanation section mark (\#\#\# Explanation), with a response (as a part of the explanation and also in the response section).

An issue with current LLMs is the limited text window it can support. This long context issue however will be addressed with future LLMs. At this stage, we test the framework with an openly available LLM, Llama 2, which supports 4,096 tokens as input, sufficient for a low or medium top-$k$ setting as 10 or 50.

\section{Experiments}

\subsection{Data Construction}

We adapt datasets MM-S14-Disease and MM-S14-CPP from the work in \cite{dong2023oet} for new concept placement in ontologies\footnote{\url{https://zenodo.org/records/10432003}}. The datasets are constructed by using two versions of SNOMED CT (2014.09 and 2017.03) with a text corpus where mentions are linked to UMLS. Then mapping between UMLS and SNOMED CT is also available in the UMLS. New mentions are therefore synthesised by considering the gap between the two versions of SNOMED CT. The edges to be inserted into the ontology for each new mention are also created, by finding the nearest parents and children for the new mention in the old version of SNOMED CT. The statistics of the dataset are displayed in Table \ref{data-statistics}.

\begin{table*}[t]
\centering
\scriptsize
\caption{Statistics for datasets for Concept Placement, for SNOMED CT (ver 20140901, ``S14'') under different categories: ``Disease'' and ``CPP'', i.e., \textit{\underline{C}linical finding}, \textit{\underline{P}rocedure}, and \textit{\underline{P}harmaceutical / biologic product}. A mention-edge pair or link (in $L$) denotes a mention (in $M$) and one of its directed edges in the KB. The mention-edge pair is complex (i.e. $L_{comp}$) when the edge involves a complex concept. Mentions are from the MedMentions dataset (``MM''). The numbers of edges are those having one hop (including leaf nodes to \texttt{NULL}) and two hops from any paths in the ontology. (Table adapted from the study \cite{dong2023oet}.)
}\label{data-statistics}
\begin{tabular}{c|l|ll}
\hline
\multicolumn{1}{l}{}                                                                                    &              & MM-S14-Disease         & MM-S14-CPP                 \\ \hline
\multirow{2}{3cm}{Ontology: \\\# all (\# complex)}                                                          & concepts     & 64,900 (824)           & 175,895 (2,718)            \\
            & edges       & 237,826 (4,997)        & 625,994 (19,401)           \\ \hline
\multirow{5}{3cm}{Corpus: \\\# $M$ / \# $L$ / \# $L_{comp}$} & train, in-KB & 11,812 / 887,840 / 917 & 34,704 / 1,398,111 / 9,475 \\
            & valid, in-KB & 4,248 / 383,457 / 203  & 11,707 / 548,295 / 4,305   \\
            \cline{2-4}
            & valid, out-of-KB & 329 / 672 / 10 & 568 / 979 / 13 \\
            & test, out-of-KB & 276 / 965 / 3 & 432 / 1,152 / 9\\ \hline
\end{tabular}
\vspace{-3.5mm}
\end{table*}

The number of edges (one-hop including leaf nodes and two-hop) is numerous, over 3.5 times of the number of concepts. This makes the task of placement into edges less tractable than entity linking into a concept for a mention.

We consider the unsupervised setting of concept placement common to the real-world scenario, which means that no mention-edge pairs for out-of-KB concepts are available for the training. This can, however, be approached using in-KB self-supervised data creation: we can see from Table \ref{data-statistics} that it is possible to generate edges for in-KB concepts; this is simply by looking at the directed parents and children of a concept in the current ontology (i.e., the older version of SNOMED CT). Thus, we use in-KB data for training and validation, and then use out-of-KB data solely for external validation and testing.

\subsection{Metrics}

We present new metrics for new concept placement, as insertion rate for any edges ($InR_{any}$) and for all edges ($InR_{all}$) predicted for mentions. Here ``any'' means that one of the gold edges is predicted for a mention, whereas ``all'' means that all of the gold edges are predicted. The metrics can be defined as below in Equation \ref{ins-rate-metrics}, where the value of $\mathbb{1}(x)$ is 1 where the statement x is true, otherwise 0, and $Z_i$ and $Y_i$ are the set of predicted edges and gold labels (or edges), resp. Also, we use the insertion rates at $k$ (i.e., $InR_{any}@k$ and $InR_{all}@k$) to denote the performance after predicting the top-$k$ edges, this measures whether the “positive” edges are ranked before the “negative” ones. We select $k$ as 1, 5, and 10, considering that terminologists can select from a few edges (as few as 10 or less) suggested by a system for updating an ontology. 
\begin{equation}
\label{ins-rate-metrics}
    InR_{any} = \frac{1}{|M|} \sum_{m_i \in M} \mathbb{1}(Z_i \cap Y_i \neq \emptyset);  \hspace{4mm} InR_{all} = \frac{1}{|M|} \sum_{m_i \in M} \mathbb{1}(Z_i \supseteq Y_i)
    \vspace{-1mm}
\end{equation}

The proposed metrics can be considered a loose version (``any'') and a strict version (``all'') of the example-based metrics \cite{Gibaja2015multilabel} for multi-label learning. The standard multi-label learning requires a complete set of gold labels, while ontologies that follow the open-world assumption are inherently incomplete (i.e., edges which are not in the gold standard may also be correct), thus the ranking-based metrics, $InR_{any}@k$ and $InR_{all}@k$, are more appropriate.

The insertion rate metrics can be used to evaluate both edge candidates and final edge selection. We also separately evaluate the insertion rate metrics for leaf edges (where the child edge is \texttt{NULL}) and non-leaf edges. 

\subsection{Experimental Settings and Baseline Methods}
We select two representative top-$k$ values after the edge enrichment step, $k=10$ and $k=50$, enriched from $\frac{k}{2}$ edges, or 5 and 25 seed edges resp., after an initial investigation of a range of $k$ values\footnote{We also investigated $k$ up to 300, while the insertion rate at $k$ improves, the overall results after edge selection are worse than smaller $k$ values as 10 and 50. A larger $k$ also leads to a substantially longer running time for edge enrichment and selection.}. Then for each of the top-$k$ settings, the models select the final set of top 1, 5, and 10 edges after the edge selection step.

For edge search, the baselines include an inverted index based approach, fixed BERT embeddings, and fine-tuned BERT embeddings with contrastive learning (as Edge-Bi-encoder). For all methods, the sub-token length of contexts and concepts are 32 and 128 resp. We choose SapBERT \cite{liu2021sap} as the BERT model in edge search (fixed and fine-tuned embeddings). For the inverted index based approach, we create an inverted index from all SNOMED CT concepts, where a key is a sub-token from a concept and a corresponding value is all the concepts, and we use the index of sub-tokens created using the SentencePiece tokenizer (also used by FLAN-T5) \cite{kudo-richardson-2018-sentencepiece}. The similarity score based on the inverted index between a mention and a concept is then calculated as the sum of inverse document frequency scores ($\text{sim}_{idf}(m,C,\mathcal{T},I,|D|) = \sum_{t \in \mathcal{T}(C) \cap \mathcal{T}(m)} \log \frac{|D|}{|I[t]|}$) of all the common sub-tokens $t$ that appear in both the mention $m$ and a concept $C$ in the set of all concepts $D$, and $\mathcal{T}$ is the tokenizer and $I$ is the index from a sub-token $t$ to the list of concepts. Then the edge score w.r.t. a mention is calculated similarly to Equation \ref{edge-emb-score-fixed}, as the average of mention-parent similarity and mention-child similarity score using the inverted index. For inverted index and fixed embedding, we use the mention only without contexts, considering that methods do not learn the relation between the concept and the natural language context; for Edge-Bi-encoder we explored mentions with or without contexts.

We apply all baseline methods with the steps of edge formation and enrichment. Then, for edge selection, we choose PubMedBERT \cite{gu2022pmbert} as the model for fine-tuning cross-encoder-based method; we also choose GPT-3.5 (``gpt-3.5-turbo-0613'')\footnote{\url{https://platform.openai.com/docs/models/gpt-3-5}}, and Llama 2 \cite{touvron2023llama} for the zero-shot prompting of LLMs, both models allowing 4,096 sub-tokens as input. FLAN-T5 \cite{chung2022instuneFlan-T5} has a limited input token length of 512, below the token usage of our prompts with the top-50 settings (between 1,556 and 3,014 sub-tokens for the datasets for top-50), thus we only use it for the top-10 setting. The model GPT-4 has a much higher cost, 30 folds of the price compared to GPT-3.5, and is slower and less stable in querying, and GPT-4 is also under updating, thus we only report results for GPT-3.5. 

For LLM instruction-tuning, we use the Supervised Fine-tuning (SFT)\footnote{\url{https://huggingface.co/docs/trl/sft\_trainer}} with 4-bit quantisation to fine-tune the Llama-2 model; the efficient instruction-tuning uses QLoRA, quantisation with Low Rank Adapters (LoRA) \cite{dettmers2023qlora}.

For all supervised models (fine-tuning and instruction-tuning), we use in-KB data for training. The best models were selected by using the validation set of the in-KB data. We then report results on the validation and the test sets for the out-of-KB data. Note that the out-of-KB validation set is not used for parameter tuning and is independent of model development.\footnote{More details on experimental settings and time usage are in Appendix 1.}

\subsection{Results}

We report results on the first two steps to determine the best edge search methods, followed by the overall results of the full framework, with edge selection. The metric results in all Tables are presented as percentage scores. 

\subsubsection{Results on Edge Search, Formation and Enrichment} Results are presented in Table \ref{results:edge-search}. The ``all'' metrics are generally lower than the ``any'' metrics (also for results in the other tables) as the full completion for concept placement is more challenging than the placement into any correct edges.

It can be observed that the fine-tuned Edge-Bi-encoder achieves the best overall results under the settings. The inverted index approach has a higher coverage of leaf edges for Diseases (but not for the broader categories of CPP) - this may be because the parent disease names are likely to be lexically similar to the new mention, while for non-leaf edges, fixed and fine-tuned embedding-based methods achieve higher performance; also, the inverted index and fixed embeddings tend to prioritise leaf edges, based on the rule by checking whether the top seed concept is a leaf concept.

\begin{table}[t]
\scriptsize
\caption{Results on edge search, formation and enrichment for MM-S14-Disease and MM-S14-CPP datasets. Each setting has validation and testing results, separated by a slash (/) sign. ``\textit{lf}'' and ``\textit{nlf}'' mean \textit{leaf} and \textit{non-leaf}, resp.
} \label{results:edge-search}
\begin{tabular}{ll|l|l|l|l|l|l}
\cline{1-8}
MM-S14-Disease                 & $k$   & $InR_{any}$   & $InR_{all}$   & $InR_{any}$, \textit{lf} & $InR_{all}$, \textit{lf} & $InR_{any}$, \textit{nlf} & $InR_{all}$, \textit{nlf} \\
\cline{1-8} 
Inverted   Index                & 10 & 10.0 / 12.0 & 9.1 / 10.1 & 9.5 / 14.2 & 9.2 / 11.9 & 14.3 / 3.4 & 8.6 / 3.4 \\
                                & 50                 & 41.3 / 40.6   & 37.7 / \textbf{38.8}   & 44.6 / \textbf{50.0}       & \textbf{41.2} / \textbf{48.2}       & 14.3 / 5.2            & 8.6 / 3.4             \\
\cline{1-8}
Fixed   embs              & 10 & 16.1 / 13.0 & 7.0 / 12.3 & 18.0 / 16.1 & 7.8 / 16.0 & 0.0 / 1.7 & 0.0 / 0.0 \\
                                & 50                 & 35.3 / 31.9   & 28.3 / 30.8   & 38.4 / 38.1       & 30.6 / 37.6       & 8.6 / 8.6             & 8.6 / 5.2             \\
\cline{1-8}
Fine-tuned embs                 & 10 & 31.9 / 25.7 & 14.6 / 8.0 & 28.9 / 12.4 & 14.6 / 8.7 & 57.1 / 75.9 & 14.3 / 5.2 \\
(Edge-Bi-enc)                                & 50        & \textbf{57.8} / \textbf{50.0}   & \textbf{40.1} / 38.0   & \textbf{55.4} / 38.1       & 38.4 / 33.5       & \textbf{77.1} / \textbf{94.8}           & \textbf{54.3} / \textbf{55.2}           \\
\hline\hline
MM-S14-CPP                 & $k$   & $InR_{any}$   & $InR_{all}$   & $InR_{any}$, \textit{lf} & $InR_{all}$, \textit{lf} & $InR_{any}$, \textit{nlf}& $InR_{all}$, \textit{nlf} \\
\cline{1-8}
Inverted   Index  & 10  & 5.5 / 5.8 & 5.1 / 5.3 & 5.3 / 5.7 & 5.1 / 5.7 & 6.9 / 6.3 & 5.2 / 3.1 \\
                                & 50           & 23.1 / 23.4 & 21.0 / 22.5 & 24.9 / 26.9 & 22.8 / 25.8 & 6.9 / 3.1   & 5.2 / 3.1   \\
\cline{1-8}
Fixed   embs     & 10 & 11.3 / 8.3 & 8.3 / 7.4 & 12.4 / 9.2 & 9.2 / 8.7 & 1.7 / 3.1 & 0.0 / 0.0 \\
                                & 50           & 28.4 / 26.9 & 25.9 / 25.2 & 30.4 / 30.2 & 28.8 / 29.6 & 10.3 / 7.8  & 0.0 / 0.0   \\
\cline{1-8}
Fine-tuned embs        & 10 & 32.0 / 27.8 & 19.7 / 14.4 & 31.2 / 19.3 & 21.4 / 16.0 & 39.7 / 76.6 & 5.2 / 4.7 \\
(Edge-Bi-enc)                                & 50           & \textbf{50.9} / \textbf{48.4} & \textbf{36.8} / \textbf{34.5} & \textbf{50.4} / \textbf{42.4} & \textbf{38.8} / \textbf{38.0} & \textbf{55.2} / \textbf{82.8} & \textbf{19.0} / \textbf{14.1} \\
\cline{1-8}
\end{tabular}

\end{table}

\begin{table}[t]
\scriptsize
\caption{Overall results after edge selection for MM-S14-Disease and MM-S14-CPP datasets. Each setting has validation and testing results, separated by a slash (/) sign.} \label{results:edge-selection}
\begin{tabular}{ll|ll|ll|ll}
\cline{1-8}                       
MM-S14-Disease   &  $k$     & $InR_{any}$@1      & $InR_{all}$@1      & $InR_{any}$@5      & $InR_{all}$@5      & $InR_{any}$@10     & $InR_{all}$@10     \\
\cline{1-8}
Inverted   Index          & 10   & 0.6 / 0.0    & 0.0 / 0.0 & 1.8 / 2.5 & 0.9 / 0.7 & 10.0 / 12.0 & 9.1 / 10.1                       \\
& 50                       & 0.6 / 0.0    & 0.0 / 0.0       & 0.9 / 1.8     & 0.0 / 0.0       & 3.3 / 4.0     & 0.9 / 1.5    \\
\cline{1-8}
Fixed   embs              & 10                       & 4.0 / 1.4 & 0.9 / 0.0 & 6.7 / 2.2 & 1.5 / 0.7 & 16.1 / 13.0 & 7.0 / \textbf{12.3} \\
& 50 & 4.0 / 1.4    & 0.9 / 0.0     & 6.7 / 2.2     & 1.5 / 0.7   & 13.4 / 4.4    & 3.0 / 2.5    \\
\cline{1-8}
Edge-Bi-enc                 & 10   & 4.0 / 11.6   & 0.3 / 0.0     & 9.7 / \textbf{17.4}    & 2.7 / 1.4   & \textbf{31.9} / 25.7 & \textbf{14.6} / 8.0                 \\
& 50 & 4.0 / 11.6   & 0.3 / 0.0     & 9.7 / \textbf{17.4}    & 2.7 / 1.4   & 13.7 / 20.3   & 4.3 / 2.5 \\
\cline{1-8}
+ Edge-Cross-enc               & 10 & 0.6 / 2.2 & 0.0 / 0.4 & 12.2 / 14.1 & 1.5 / 3.6 & \textbf{31.9} / 25.7 &  \textbf{14.6} / 8.0   \\
& 50                 & \textbf{7.3} / 7.6    & \textbf{1.8} / \textbf{1.5}   & \textbf{17.9} / 15.6   & \textbf{7.3} / \textbf{4.7}   & 25.8 / \textbf{26.5}   & 10.6 / 8.7   \\
\cline{2-8}
+ GPT-3.5                       & 10 & 4.0 / 4.0 & 0.0 / 0.0 & 5.5 / 4.3 & 2.4 / 1.4 & 5.5 / 4.3 & 2.4 / 1.4 \\
                                & 50                & 3.3 / 1.5  & 0.0 / 0.0       & 4.6 / 3.6   & 1.5 / 0.4  & 4.6 / 3.6   & 1.5 / 0.4     \\
\cline{2-8}
+ FLAN-T5-XL                      & 10 & 2.7 / 1.8 & 0.6 / 0.0 & 2.7 / 1.8 & 0.6 / 0.0 & 2.7 / 1.8 & 0.6 / 0.0 \\
\cline{2-8}
+ Llama-2-7B                      
                                & 10 & 2.7 / 4.3 & 0.3 / 0.0 & 5.8 / 6.2 & 2.1 / 0.0 & 8.8 / 7.2 & 3.3 / 1.1 \\
                                & 50                 & 1.8 / 3.3  & 0.0 / 0.0       & 3.7 / 5.8   & 1.2 / 0.7  & 4.0 / 6.9   & 1.2 / 0.7    \\
\cline{2-8}
+ Llama-2-7B-tuned                   & 10  & 5.2 / \textbf{13.8} & 0.0 / 0.0 & 7.6 / 16.3 & 1.5 / 1.8 & 7.6 / 16.3 & 1.5 / 1.8\\
                                & 50                 & 6.1 / 13.0 & 0.0 / 0.0       & 8.5 / 15.2  & 1.5 / 1.1  & 8.5 / 15.6  & 1.5 / 1.5               \\
\hline\hline
MM-S14-CPP   &  $k$     & $InR_{any}$@1      & $InR_{all}$@1      & $InR_{any}$@5      & $InR_{all}$@5      & $InR_{any}$@10     & $InR_{all}$@10     \\
\cline{1-8}
Inverted   Index          & 10 & 0.4 / 0.0   & 0.0 / 0.0   & 0.9 / 0.5     & 0.0 / 0.0 & 5.5 / 5.8 & 5.1 / 5.3                             \\
& 50 & 0.4 / 0.0   & 0.0 / 0.0   & 0.4 / 0.0     & 0.0 / 0.0   & 0.9 / 1.4     & 0.0 / 0.5\\
\cline{1-8}
Fixed   embs              
& 10 & 2.8 / 1.2 & 0.7 / 0.2 & 6.3 / 3.7 & 2.3 / 2.1 & 11.3 / 8.3 & 8.3 / 7.4 \\
& 50                        & 2.8 / 1.2   & 0.7 / 0.2   & 6.3 / 3.7     & 2.3 / 2.1   & 7.9 / 6.0     & 3.9 / 4.6     \\
\cline{1-8}
Edge-Bi-enc                  & 10   & 2.5 / 6.3   & 0.0 / 0.2   & 6.2 / 11.8    & 1.2 / 1.9   & \textbf{32.0} / \textbf{27.8} & \textbf{19.7} / \textbf{14.4}                        \\
& 50 & 2.5 / 6.3   & 0.0 / 0.2   & 6.2 / 11.8    & 1.2 / 1.9   & 8.6 / 14.4    & 3.0 / 3.5 \\
\cline{1-8}
+ Edge-Cross-enc            & 10 & 3.4 / \textbf{9.3} & 0.2 / 0.0  & 7.8 / 13.7 & 2.1 / 2.3  & \textbf{32.0} / \textbf{27.8} & \textbf{19.7} / \textbf{14.4}  \\
                            & 50                 & 4.9 / 3.9   & \textbf{2.1} / 0.2   & \textbf{15.3} / \textbf{17.6}   & \textbf{6.3} / \textbf{6.9}   & 24.8 / 26.6   & 13.2 / 14.4   \\
\cline{2-8}
+ GPT-3.5                   & 10 & \textbf{5.1} / 3.9 & 0.0 / 0.0 & 7.9 / 6.0 & 3.3 / 3.5 & 7.9 / 6.0 & 3.3 / 3.4\\
                            & 50    & 1.8 / 1.9 & 0.0 / 0.0 & 3.9 / 2.8 & 0.9 / 0.7 & 4.0 / 2.8 & 0.9 / 0.7                  \\
\cline{2-8}
+ FLAN-T5-XL                      & 10 & 2.6 / 1.9 & 0.5 / \textbf{0.7} & 2.6 / 1.9 & 0.5 / 0.7 & 2.6 / 1.9 & 0.5 / 0.7 \\
\cline{2-8}
+ Llama-2-7B                      & 10 & 1.8 / 4.6 & 0.0 / 0.2 & 4.8 / 7.2 & 0.7 / 1.9 & 8.8 / 10.4 & 3.9 / 3.9 \\
& 50 &  1.2 / 3.5	& 0.0 / 0.0	& 2.5 / 5.1 & 	0.7 / 0.9	& 3.0 / 6.3	& 1.1 / 1.2 \\
\cline{2-8}
+ Llama-2-7B-tuned                   & 10 & 2.6 / 7.2 & 0.0 / 0.0 & 6.5 / 10.6 & 1.9 / 1.2 & 7.6 / 12.7 & 2.5 / 3.2 \\
& 50        & 2.5 / 4.6 & 0.0 / 0.0 & 3.3 / 6.7 & 0.5 / 0.6 & 4.0 / 8.1 & 0.9 / 1.4               \\
\cline{1-8}
\end{tabular}
\vspace{-3.5mm}
\end{table}

\subsubsection{Overall Results after Edge Selection}

We then add the edge selection steps mainly on the candidates from the fine-tuned embedding (Edge-Bi-encoder) approach, given its best overall performance in generating edge candidates. 

As shown in Table \ref{results:edge-selection}, the multi-label Edge-Cross-encoder achieves the best performance in most experimental settings. Edge-Cross-encoder further reranks the edge candidates and helps substantially improve the performance over Edge-Bi-encoder (edge search only), e.g., by around 8-9\% absolute scores for $InR_{any}@5$ for the datasets and around 12-16\% for $InR_{any}@10$ (except for the same @10 results for top-10 setting, where re-ranking does not make a difference).

We also test the LLMs, it can be seen that the tested medium scale LLMs (GPT-3.5, FLAN-XL, and Llama-2-7B), especially not instruction-tuned for the task, can still not be directly used for concept placement, although GPT-3.5 has notably better results on top edge suggestion ($InR_{any}@1$) from the top-10 setting. The explainable instruction tuning approach greatly improves the performance of Llama-2-7B. This shows that training on in-KB data by generating an automated explanation before generating the results is practically useful to enhance the capability of LLMs on ontology reasoning tasks. Most results from LLMs, except for the top edge suggestion ($InR_{any}@1$ and $InR_{all}@1$), are still below the original candidates from Edge-Bi-encoder. Nevertheless, the results are encouraging and can motivate future studies using LLMs for concept placement.

We also notice a performance gap between the validation set and the test set on the two datasets in Tables \ref{results:edge-search}-\ref{results:edge-selection}, which may be due to the high variance caused by the small number of mentions in the sets (between 200 and 600, see Table \ref{data-statistics}) and the distinct data distribution based on concept drift (e.g., different lexical mentions between the sets), showing the challenge to generalise to new concepts.

\vspace{-3.5mm}

\subsubsection{Discussion on Model Applicability} The overall performance of the models is not high, especially for LLMs, as shown in Table \ref{results:edge-selection}. The best InAny@10 is around 30\% with Edge-Bi-encoder and Edge-Cross-encoder. This shows that the models cannot support an automated application, but still, they may potentially be applicable to suggest a ranking of the edges for human terminologists to add a new concept to an ontology. In practice, having a larger $k$ can help improve the metrics of $InR_{any}@k$ and $InR_{all}@k$, but can also increase the effort of manual selection, thus a balance needs to be achieved and warrants future studies.

\vspace{-3.5mm}

\subsubsection{Case Study} In Appendix 3, we select a few test mentions and display the 5 top edge suggestions, under the top-50 setting from Edge-Bi-encoder, Edge-Cross-encoder, Llama-2-7B, and Llama-2-7B fine-tuned models. For Llama-2 models, we display the generated answers. Without instruction tuning, Llama-2-7B sometimes generates answers in an incorrect format or generates irrelevant outputs. With explainable instruction-tuning, Llama-2-7B generates explanations that follow a natural language reasoning path to lead to the correct edge option. We also note that merging with existing concepts is needed as a further step after the placement of the mention, e.g., for Chronic kidney disorder.

We also note that many edge predictions are not completely wrong, for example for the first case, the predicted parent (e.g., kidney disease)  in the methods is more general than the gold direct parent (e.g., renal impairment), and the predicted child (e.g., hypertensive heart and renal disease with renal failure) is not far from the gold children in the ontology structure. However, calculating a lenient, soft score (e.g., with Wu \& Palmer similarity \cite{wu-palmer-1994-verb}) between every prediction and the set of gold edges instead of a binary evaluation is not time efficient in our experiments. 
We leave an efficient, lenient evaluation for future studies.

\subsection{Ablation Studies} 

\begin{figure}[t]
\centering
  \includegraphics[width=0.49\textwidth]{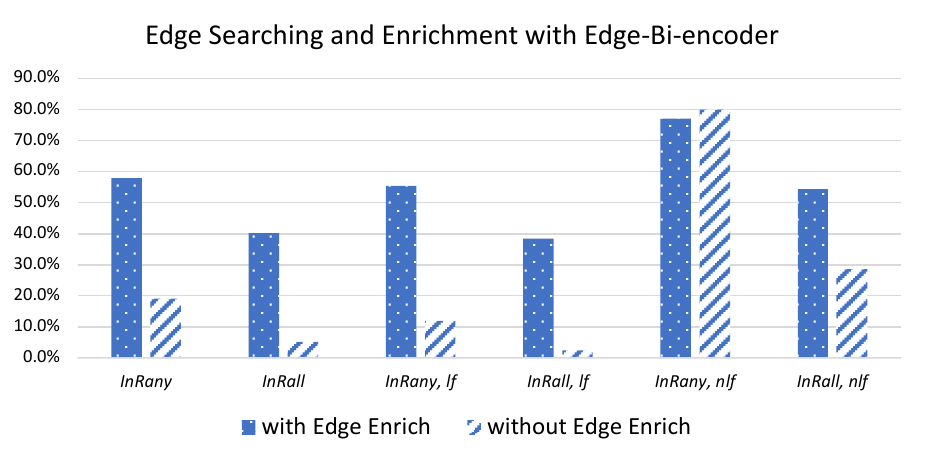}
  \includegraphics[width=0.49\textwidth]{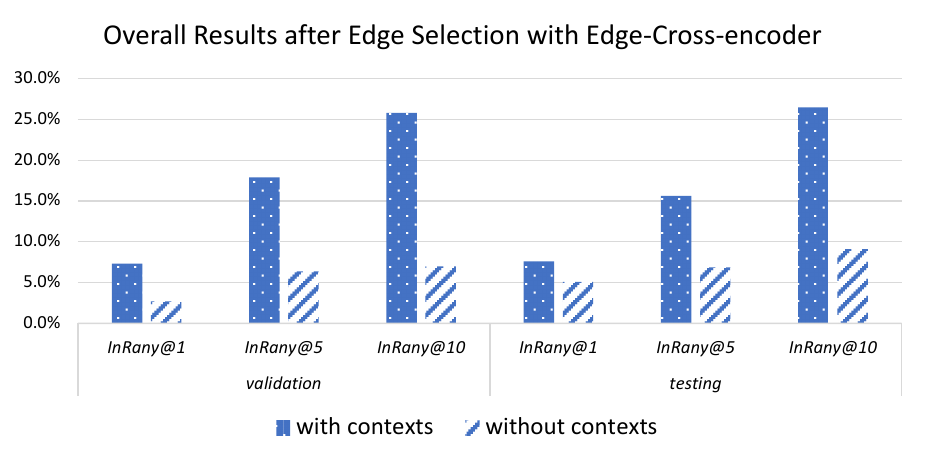}
  \includegraphics[width=0.57\textwidth]{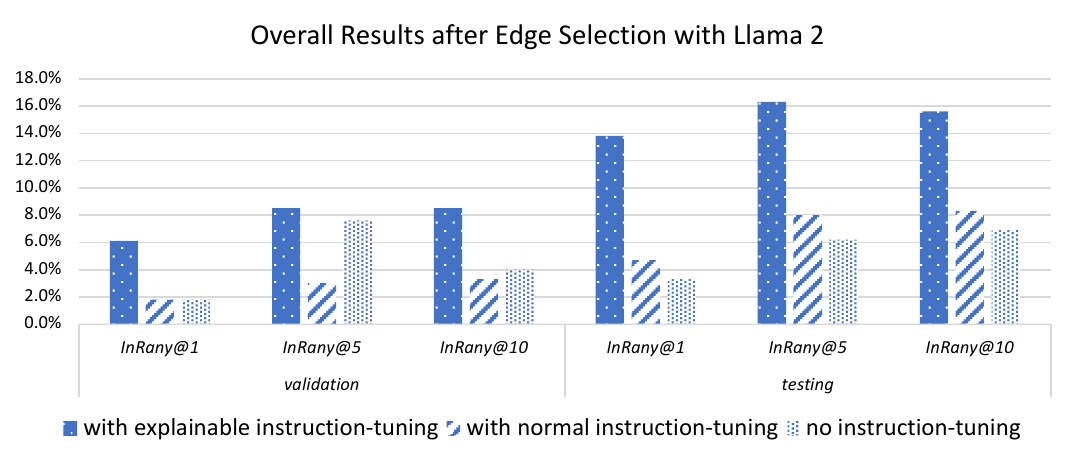}
  \caption{Ablation results on top-50 edge candidates with MM-S14-Disease dataset: \textbf{(a)} Top-left: results with Edge-Bi-encoder, with or without the edge enrichment step, on validation set; \textbf{(b)} Top-right: overall results with Edge-Cross-encoder, with or without contexts; \textbf{(c)} Bottom: overall results after Edge Selection with Llama-2-7B, with explainable instruction-tuning, normal instruction-tuning, or without instruction-tuning.}\label{Figs:ablation}
  
\end{figure}

Our ablation studies aim to investigate how edge enrichment and automated explainable instruction tuning can enhance the performance of concept placement. We use top-50 setting and MM-S14-Disease dataset as an example, and other top-$k$ settings and MM-S14-CPP dataset follow a similar pattern of results.

\vspace{-3.5mm}

\subsubsection{Edge Enrichment} Edge enrichment has greatly improved the overall results for Edge-Bi-encoder, about above 30\% absolute improvement of $InR_{any}$ and $InR_{all}$ as shown in Figure \ref{Figs:ablation} part (a). Results with inverted index and fixed embeddings are in Appendix 2, showing a general improvement with edge enrichment. 

\vspace{-3.5mm}

\subsubsection{Contextual Information of Mention} The left and right contexts of a mention are useful in edge search with Edge-Bi-encoder to learn the similarity between a contextual mention and a concept, even though the type of texts is distinct from ontology concept labels, as shown in Figure \ref{Figs:ablation} part (b). 

\vspace{-3.5mm}

\subsubsection{Explainable Instruction Tuning} Explainable instruction tuning helps improve the performance of the LLM, Llama-2-7B, under the top-50 setting, especially on the test set of MM-S14-Disease, as displayed in Figure \ref{Figs:ablation} part (c). In contrast, normal instruction tuning that directly generates the edge option number does not always improve over the case without instruction tuning. 

\subsection{Conclusion and Future Studies}

We propose an LM-based framework for new concept placement in ontologies. The framework uses a three-step approach, that enhances the two-step information retrieval with edge formation and enrichment leveraging the ontological structure. The results overall show that methods that fine-tune PLMs perform the best, while there is an encouraging performance for the recent LLMs, especially with explainable instruction tuning. Our case study shows that explanations can be generated to detail the steps for concept placement. We focused on placing directly the mentions into edges in this work, a following step is to group or merge mentions of the same new concept, and also with existing concepts if they have the same meaning, when they are placed into the same edges. Future studies will further explore LLM-generated explanations and leverage advanced Retrieval Augmented Generation \cite{gao2023rag-survey} and prompting strategies. Future studies also need to investigate how to use the methods to assist human terminologists.

\noindent\textbf{Acknowledgements}. This work is supported by EPSRC projects, including ConCur (EP/V050869/1), OASIS (EP/S032347/1), UK FIRES (EP/S019111/1); and Samsung Research UK (SRUK).

\bibliographystyle{splncs04}
\bibliography{references}

\clearpage\section*{Appendix 1: Detailed model settings and time usage}

The approaches are implemented using PyTorch and Huggingface Transformers. Edge-Bi-encoder and Edge-Cross-encoder are originally based on the architectures of BLINKout \cite{dong2023NIL} (based on BLINK \cite{wu2020blink}). Inverted index with ontology concepts is based on DeepOnto Library \cite{he2023deeponto}. The batch sizes for Edge-Bi-encoder and Edge-Cross-encoder are 16 and 1, resp. The fine-tuning of Edge-Bi-encoder and Edge-Cross-encoder takes 1 and 4 epochs, resp. We limit the rows to 200,000 for training the Edge-Cross-encoder models given the sufficient amount the data for model convergence and the long time of training. The instruction tuning of Llama-2-7B uses a 4-bit quantisation and takes 3 epochs with a batch size of 4. 

\subsubsection*{Time usage} We run all models using an NVIDIA Quadro RTX 8000 GPU card (48GB GPU). We report the time usage estimate for MM-S14-Disease under the top-50 setting. Training bi-encoder took around 29 hours. Training cross-encoder took around 4 hours. Instruction tuning of Llama-2-7B took around 16 hours. Inferencing with fixed embeddings and inverted index with edge enrichment is within around 0.5 and 1 second per mention, resp. Inferencing with Edge-Bi-encoder only takes around 0.2 second per mention. The whole inferencing with both Edge-Bi-encoder and Edge-Cross-encoder takes around 2.3 seconds per mention. The prompting of an explainable instruction-tuned Llama-2-7B model takes around 78 seconds per mention to output natural language explanations.

\section*{Appendix 2: Detailed results on edge enrichment}

We applied edge formation enrichment over inverted index and fixed embedding approach. Results in Table \ref{results:edge-search-enrich} show a substantial improvement for $InR_{any}$ and $InR_{all}$. We see that the mentions to be placed to non-leaf edges are not improved with inverted index and fixed embeddings, but are improved with the fine-tuned, Edge-Bi-encoder, this is because the latter places a more lenient score for the leaf edges that do not always rank them before the non-leaf edges. 

\begin{table}[h]
\scriptsize
\caption{Results on edge search and enrichment (vs. not using edge enrichment) for MM-S14-Disease, under the top-50 setting. Each setting has validation and testing results, separated by a slash (/) sign. ``\textit{lf}'' and ``\textit{nlf}'' mean \textit{leaf} and \textit{non-leaf}, resp.} \label{results:edge-search-enrich}
\begin{tabular}{ll|l|l|l|l|l|l}
\cline{1-8}
MM-S14-Disease                 & $k$   & $InR_{any}$   & $InR_{all}$   & $InR_{any}$, \textit{lf} & $InR_{all}$, \textit{lf} & $InR_{any}$, \textit{nlf} & $InR_{all}$, \textit{nlf} \\
\cline{1-8} 
Inverted   Index          & 50                 & \textbf{41.3} / \textbf{40.6}   & \textbf{37.7} / \textbf{38.8}   & \textbf{44.6} / \textbf{50.0}       & \textbf{41.2} / \textbf{48.2}       & 14.3 / 5.2            & 8.6 / 3.5             \\
w/o Edge Enrich         & 50                 & 7.3 / 6.5 & 4.6 / 3.6 & 2.7 / 3.7 & 0.7 / 0.5 & \textbf{45.7} / \textbf{17.2} & \textbf{37.1} / \textbf{15.5}             \\
\cline{1-8}
Fixed   embs              & 50                 & \textbf{35.3} / \textbf{31.9}   & \textbf{28.3} / \textbf{30.8}   & \textbf{38.4} / \textbf{38.1}       & \textbf{30.6} / \textbf{37.6}       & 8.6 / 8.6             & \textbf{8.6} / \textbf{5.2}             \\
w/o Edge Enrich         & 50                 & 19.1 / 17.3       & 4.3 / 4.3 & 16.0 / 6.4 & 3.7 / 4.1 & \textbf{45.7} / \textbf{58.6} & \textbf{8.6} / \textbf{5.2}      \\
\cline{1-8}
Edge-Bi-Enc                 & 50        & \textbf{57.8} / \textbf{50.0}   & \textbf{40.1} / \textbf{38.0}   & \textbf{55.4} / \textbf{38.1}      & \textbf{38.4} / \textbf{33.5}       & 77.1 / \textbf{94.8}           & \textbf{54.3} / \textbf{55.2}           \\
w/o Edge Enrich         & 50         & 19.1 / 23.9 & 5.2 / 5.8 &  11.9 / 5.0 & 2.4 / 2.8 & \textbf{80.0} / \textbf{94.8} & 28.6 / 17.2                   \\
\cline{1-8}
\end{tabular}
\end{table}

\vspace{-8mm}

\section*{Appendix 3: Qualitative examples}

Examples of a non-leaf and a leaf concept placement, with prompt options, model predictions, and instruction-tuned Llama-2-7B's explanations, are in Table \ref{case-study}. 

\begin{table}[t]
\tiny
\caption{Examples of two mentions in the out-of-KB test set of MM-S14-Disease to enrich SNOMED CT 2014.09. The correct predictions are in \textbf{bold}. (Note: while the concept Chronic kidney disease in SNOMED CT ver 2017.03 is not in ver 2014.09, it is modified from Chronic renal impairment, ID 236425005, in the older ontology.)}\label{case-study}
\begin{tabular}{p{1.5cm}|p{5.3cm}|p{5.1cm}}
\cline{1-3}
 & \multicolumn{1}{c}{Test, out-of-KB, 13} & \multicolumn{1}{|c}{Test, out-of-KB, 138} \\
 \cline{1-3}
*\textbf{Mention}* in contexts                    & ...Since no one had *\textbf{CKD}* in   partial nephrectomized patients, we determined risk factors for CKD in   radical nephrectomized patients…                                                                                                                                                                                                                                                                                                                                                                                                                                                                                                                                                                                                                                                                                                                                                                                                                                                                                                                              & Development of a novel   near-infrared fluorescent theranostic combretastain A-4 analogue, YK-5-252,   to target triple negative breast cancer. The treatment of triple negative   breast cancer (*\textbf{TNBC}*) is a significant challenge to cancer research...                                                                                                                                                                                                                                                                                                                                                                                                                                                                                                                        \\
\cline{1-3}
Gold Concept & \begin{tabular}[c]{@{}p{5.3cm}@{}}\url{http://snomed.info/id/709044004} \\Chronic kidney disease \\ (not available in SNOMED CT 2014.09) \end{tabular}   & \begin{tabular}[c]{@{}p{5.1cm}@{}}\url{http://snomed.info/id/706970001} \\  Triple negative malignant neoplasm of breast \\ (not available in SNOMED CT 2014.09) \end{tabular} \\ 
\cline{1-3}
Gold Edges                        & \begin{tabular}[c]{@{}p{5.3cm}@{}}\textbf{Parents}: (i) Renal impairment  $\rightarrow$ \\\textbf{Children}: (i) Chronic renal   impairment associated with type II diabetes mellitus; (ii) Hypertensive heart and   chronic kidney disease; (iii) Chronic kidney disease stage 1; (iv) Chronic kidney   disease stage 2; (v) Chronic kidney disease stage 3; (vi) Chronic kidney disease stage   4; (vii) Chronic kidney disease stage 5; (viii) Chronic renal failure syndrome; (viiii) Hypertensive heart AND chronic kidney   disease on dialysis; (x) Chronic kidney disease due to hypertension; Malignant   hypertensive chronic kidney disease \end{tabular}                                                                                                                                                                                                                                                                                                                                                                                                                                                                                                                              & \begin{tabular}[c]{@{}p{5.1cm}@{}}\textbf{Parents}: (i) Human epidermal growth   factor 2 negative carcinoma of breast; (ii) Malignant tumor of breast; (iii) Hormone   receptor negative neoplasm  $\rightarrow$ \\\textbf{Children}: (i) NULL  \end{tabular}                                                                                                                                                                                                                                                                                                                                                                                                                                                                                                                                                                                                                    \\
\cline{1-3}
Edge-Bi-enc                  & \begin{tabular}[c]{@{}p{5.3cm}@{}}(i) renal impairment  $\rightarrow$ end stage renal disease \\      (ii) renal impairment  $\rightarrow$ renal   failure following molar and/or ectopic pregnancy \\      (iii) renal impairment  $\rightarrow$ renal   failure syndrome \\      \textbf{(vi) renal impairment  $\rightarrow$ chronic   kidney disease due to hypertension} \\      (v) kidney disease  $\rightarrow$ impaired   renal function disorder\end{tabular}                                                                                                                                                                                                                                                                                                                                                                                                                                                                                                                                                                                                                             & \begin{tabular}[c]{@{}p{5.1cm}@{}}(i) malignant tumor of breast  $\rightarrow$ lobular carcinoma of breast \\  (ii)    carcinoma of breast  $\rightarrow$ lobular   carcinoma of breast \\     (iii) malignant tumor of breast  $\rightarrow$   mucinous carcinoma of breast \\   (iv)   carcinoma of breast  $\rightarrow$ cancer   en cuirasse \\  (v)    malignant tumor of breast  $\rightarrow$   malignant phyllodes tumor of breast\end{tabular}                                                                                                                                                                                                                                                                                                                                            \\
\cline{1-3}
Edge-Cross-enc               & \begin{tabular}[c]{@{}p{5.3cm}@{}}(i) kidney disease  $\rightarrow$ renal function   impairment with growth failure \\ (ii) kidney disease  $\rightarrow$ impaired renal function disorder \\ (iii) disorder of the genitourinary system    $\rightarrow$ impaired renal function disorder \\ \textbf{(iv) renal impairment  $\rightarrow$ chronic kidney disease due to   hypertension} \\ (v) kidney disease  $\rightarrow$   renal failure syndrome  \end{tabular}

                            & \begin{tabular}[c]{@{}p{5.1cm}@{}}(i) neoplasm of breast  $\rightarrow$ malignant   phyllodes tumor of breast\\ (ii) carcinoma of breast  $\rightarrow$ cancer en cuirasse  \\ \textbf{(iii) malignant tumor of breast  $\rightarrow$ NULL}\\ (iv) carcinoma of breast  $\rightarrow$ NULL\\ (v) neoplasm of breast  $\rightarrow$ NULL \end{tabular}\\
\cline{1-3}
Prompt, only options are displayed (correct options in \textbf{bold}), the full prompt template is in Sect \ref{sec:edge-select}                      & \begin{tabular}[c]{@{}p{5.3cm}@{}}options:\\      0.renal impairment  $\rightarrow$ end stage   renal disease \\      1.renal impairment  $\rightarrow$ renal   failure following molar and/or ectopic pregnancy \\      2.renal impairment  $\rightarrow$ renal   failure syndrome \\      \textbf{3.renal impairment  $\rightarrow$ chronic   kidney disease due to hypertension} \\      4.kidney disease  $\rightarrow$ impaired   renal function disorder \\      5.kidney disease  $\rightarrow$ renal   function impairment with growth failure \\      6.kidney disease  $\rightarrow$ renal failure   syndrome \\      \textbf{7.renal impairment  $\rightarrow$ chronic   renal failure syndrome} \\      8.disorder of the genitourinary system    $\rightarrow$ impaired renal function disorder \\      9.renal impairment  $\rightarrow$   hypertensive heart and renal disease with renal failure\end{tabular}                                                                                                                                                                  & \begin{tabular}[c]{@{}p{5.1cm}@{}}options:\\      0.malignant tumor of breast  $\rightarrow$   lobular carcinoma of breast \\      1.carcinoma of breast  $\rightarrow$ lobular   carcinoma of breast \\      2.malignant tumor of breast  $\rightarrow$   mucinous carcinoma of breast \\      3.carcinoma of breast  $\rightarrow$ cancer   en cuirasse \\      4.malignant tumor of breast  $\rightarrow$   malignant phyllodes tumor of breast \\      5.malignant tumor of breast  $\rightarrow$   cancer en cuirasse \\      6.neoplasm of breast  $\rightarrow$ malignant   phyllodes tumor of breast \\      \textbf{7.malignant tumor of breast  $\rightarrow$   NULL}\\      8.carcinoma of breast  $\rightarrow$   NULL\\      9.neoplasm of breast  $\rightarrow$   NULL\end{tabular} \\
\cline{1-3}
Gold answer  &    3,7 (option incomplete) & 7 (option incomplete) \\
\cline{1-3}
Llama-2-7B                   & 1 & Fail to generate answer with the   relevant content\\
\cline{1-3}
Llama-2-7B instruction-tuned & \begin{tabular}[c]{@{}p{5.3cm}@{}}\#\#\# Explanation:\\      From the parents in the options above, including renal impairment , kidney   disease , disorder of the genitourinary system , the correct parents of the   mention, CKD, include \textbf{renal impairment} , kidney disease , disorder of the   genitourinary system . Thus the options are narrowed down to 0, 1, 2, 3, 7,   9, 4, 5, 6, 8. From the children in the narrowed options, including end stage   renal disease , renal failure following molar and/or ectopic pregnancy ,   renal failure syndrome , chronic kidney disease due to hypertension , chronic   renal failure syndrome , impaired renal function disorder , renal function   impairment with growth failure , the correct children of the mention, CKD,   include \textbf{chronic kidney disease due to hypertension} , \textbf{chronic renal failure   syndrome} , impaired renal function disorder , renal function impairment with   growth failure . Thus, the final answers are \textbf{3}, \textbf{7}, 5, 6, 8.\\      \\      \#\#\# Response:\\      3,7,5,6,8\end{tabular} & \begin{tabular}[c]{@{}p{5.1cm}@{}}\#\#\# Explanation:\\      From the parents in the options above, including malignant tumor of breast   , carcinoma of breast , neoplasm of breast , the correct parents of the   mention, TNBC, include \textbf{malignant tumor of breast} . Thus the options are   narrowed down to 0, 2, 4, 5, 7. From the children in the narrowed options,   including lobular carcinoma of breast , mucinous carcinoma of breast ,   malignant phyllodes tumor of breast , cancer en cuirasse , \textbf{NULL}, the correct   children of the mention, TNBC, include NULL. Thus, the final answers are   \textbf{7}.\\      \\      \#\#\# Response:\\      \textbf{7}\end{tabular}                           \\
\cline{1-3}
\end{tabular}
\end{table}
\end{document}